# Predicting Healthcare System Visitation Flow by Integrating Hospital Attributes and Population Socioeconomics with Human Mobility Data

Binbin Lin[a], Lei Zou [a]*, Hao Tian[a], Heng Cai[a], Yifan Yang[a], Bing Zhou[b]

[a]Department of Geography, Texas A&M University, College Station, TX, USA;

[b]Department of Geography, University of Tennessee, Knoxville, TN, USA

*Email: lzou@tamu.edu

*Address: CSA 205D, 3147 TAMU, College Station, TX 77843-3147

## Acknowledgments

This study is based on work supported by the Data Resource Develop Program Award from the Texas A&M Institute of Data Science (TAMIDS). Any opinions, findings, and conclusions, or recommendations expressed in this material are those of the authors and do not necessarily reflect the views of the funding agencies.

## Data Availability Statement

The mobility data used in this study were provided by SafeGraph, Inc. through its Data for Academics program. Due to data licensing restrictions, the original datasets are not publicly available. However, qualified academic researchers can request access from SafeGraph at https://www.safegraph.com/academics.

## Disclosure statement

No potential conflict of interest was reported by the authors.

# Predicting Healthcare System Visitation Flow by Integrating Hospital Attributes and Population Socioeconomics with Human Mobility Data

Abstract: Healthcare visitation patterns are influenced by a complex interplay of hospital attributes, population socioeconomics (SES), and spatial factors. However, existing research often adopts a fragmented approach, examining these determinants in isolation. This study addresses this gap by integrating hospital capacities, occupancy rates, reputation, and popularity with population SES and spatial mobility patterns to predict visitation flows and analyze influencing factors. Utilizing four years (2020–2023) of SafeGraph mobility data and user experience data from Google Maps Reviews, five flow prediction models—Naïve Regression, Gradient Boosting, Multilayer Perceptrons (MLPs), Deep Gravity, and Heterogeneous Graph Neural Networks (HGNN)—were trained and applied to simulate visitation flows in Houston, Texas, U.S. The Shapley additive explanation (SHAP) analysis and the Partial Dependence Plot (PDP) method were employed to examine the combined impacts of different factors on visitation patterns. The findings reveal that Deep Gravity outperformed other models. Hospital capacities, ICU occupancy rates, ratings, and popularity significantly influence visitation patterns, with their effects varying across different travel distances. Short-distance visits are primarily driven by convenience, whereas long-distance visits are influenced by hospital ratings. White-majority areas exhibited lower sensitivity to hospital ratings for short-distance visits, while Asian populations and those with higher education levels prioritized hospital rating in their visitation decisions. SES further influence these patterns, as areas with higher proportions of Hispanic, Black, under-18, and over-65 populations tend to have more frequent hospital visits, potentially reflecting greater healthcare needs or limited access to alternative medical services. These findings offer actionable insights for policymakers to design targeted interventions promoting equitable healthcare access.

## 1. Introduction

Despite advancements in healthcare accessibility, significant inequality persists due to socioeconomic factors, geographic distribution, and resource allocation, disproportionately affecting vulnerable groups and exacerbating health disparities (Chen et al., 2023; McMaughan et al., 2020). Here, healthcare accessibility refers to both the spatial ease and social opportunity of obtaining medical services, encompassing geographic proximity, travel effort, and individuals' socioeconomic capacity to utilize care (Guagliardo, 2004). Healthcare inequality, in turn, denotes uneven opportunities to access and use healthcare across population groups, often arising from these spatial and social disparities.

Research on hospital accessibility has sought to address these inequalities, initially focusing on basic factors such as hospital capacity and population within service areas (D.-H. Yang et al., 2006). Subsequent research introduced factors like distance decay (Luo & Qi, 2009), reflecting the reduced likelihood of healthcare utilization as travel distances increase, and hospital quality (Tao et al., 2020), which influences patient preferences. Demographic differences are also considered to account for varying healthcare needs across populations. However, many previous studies adopt a one-dimensional approach, focusing on either hospital attributes or community demographics. In reality, healthcare access is shaped by the interplay of multiple factors, including hospital quality, capacity, proximity, and socioeconomic status (SES). For example, low-income populations may prioritize proximity due to financial constraints, while higher-income groups may favor hospitals offering superior services. This fragmented focus often limits insights into the complex dynamics of accessibility, highlighting the need for a comprehensive framework that integrates hospital and demographic factors to better address healthcare inequalities.

Understanding the compounding effects of hospital and community attributes requires robust, real-world data on healthcare access patterns. Advances in location tracking technologies have enabled detailed analysis of human mobility patterns, including healthcare accessibility and visitations. SafeGraph, a large-scale mobility data provider that aggregates anonymized location information from mobile devices, offers origin-destination (OD) datasets capturing population movement patterns across places and time (Advan Research, 2025). Among these, the healthcare-related OD flows, movements of residents from their neighborhoods to hospitals, provide a granular view of spatial healthcare utilization. In this study, we focus on predicting hospital visitation flows, which reflect the realized outcomes of healthcare accessibility shaped by actual health needs, thereby representing the true utilization of healthcare services (J. Wang et al., 2021). In addition, user-generated content from platforms such as Google Maps provides patient-centered perspectives on hospital quality, with ratings and comments reflecting visitor experiences, while the volume of reviews often serves as a proxy for hospital popularity. The integration of these novel data enables a comprehensive analysis of how hospital attributes and SES collectively influence healthcare access, addressing key limitations in traditional accessibility research.

Mobility pattern modeling has become a vibrant field of research, playing a key role in optimizing resources and improving service accessibility. Various methods, including multilayer perceptron (MLPs) (Zhang et al., 2024), deep gravity models (Simini et al., 2021), and heterogeneous graph neural networks (HGNNs) (Terroso Sáenz et al., 2023), have been developed to advance this field by predicting facility visitation flows. These methods have been applied across various contexts, including museum visitor flows, tourist foot traffic, transportation flows, and inter-city movement patterns. However, the relative effectiveness of these models in predicting healthcare facility visits remains uncertain, particularly when accounting for the interplay of hospital quality, population demographics, and spatial mobility patterns. Identifying the most suitable modeling framework is crucial for advancing healthcare accessibility research and addressing healthcare inequities.

In order to examine the joint influence of hospital attributes and neighborhood characteristics on access patterns, this study analyzed four years (2020-2023) SafeGraph human mobility data in Houston, Texas in the U.S. Key hospital factors considered include hospital capacity, occupancy, service quality, and popularity, while community factors include SES such as income, race, and educational attainment. In addition, the study incorporated distance decay effects by measuring the drive time between hospitals and census block groups. By comparing five commonly used flow prediction models, this study aims to address a central question: How do healthcare system attributes and resident characteristics shape access patterns? To explore this question, this study establishes three main objectives: (1) to integrate social sensing data (e.g., Google Maps reviews and human mobility data) into healthcare system visit prediction; (2) to identify and evaluate advanced machine learning models for predicting healthcare visits; and (3) to investigate how healthcare system attributes and community-level SES interact to shape hospital visitation patterns. The study hypothesizes that healthcare visitation patterns emerge from the joint and interacting influences of distance attenuation, hospital attributes, and community socioeconomic status (SES), rather than from any single factor in isolation. By integrating multi-source data and comparing model architectures, this study establishes an empirical framework for predicting hospital visitation flows and revealing the structural determinants of inequality in healthcare utilization. This approach provides a scalable pathway for identifying vulnerable areas and informing equity-focused, resilience-oriented health system planning.

## 2. Related work

**2.1 Factors Influencing Healthcare System Visitation**

Understanding the factors influencing healthcare system visitation is crucial for enhancing accessibility and promoting equity in healthcare resources. A well-established theoretical framework for understanding the determinants of healthcare utilization is Andersen's Behavioral Model of Health Services Use. Originally developed by Ronald M. Andersen in 1968 to explain family use of medical services, the model proposed that healthcare utilization is shaped by three key dimensions: predisposing factors (e.g., age, gender, education, social structure, and health beliefs), enabling factors (e.g., income, insurance coverage, and spatial accessibility), and need factors (both perceived and evaluated health status) (R. Andersen, 1974). The model has since undergone several major revisions. The second generation (R. Andersen & Aday, 1978) expanded the framework from the individual to the societal level, integrating system-level accessibility conditions. The third version (R. M. Andersen, 1995) introduced feedback mechanisms among health behaviors, outcomes, and healthcare systems, establishing a dynamic and multidimensional structure. Later refinements (2000-2008) incorporated contextual influences such as policy environment, community characteristics, and healthcare infrastructure, allowing applications across diverse populations and spatial scales (Alkhawaldeh et al., 2023; Babitsch et al., 2012). This conceptual foundation helps contextualize variations in hospital choice and healthcare visitation as outcomes of interacting behavioral, structural, and spatial determinants.

Building on this conceptual foundation, previous studies have identified two major groups of determinants corresponding to Andersen's enabling and predisposing components. Hospital-related factors, e.g., medical quality, reputation, patient experience, insurance coverage, hospital size, and proximity, represent enabling resources that facilitate or constrain healthcare access. The relative importance of these factors varies across institutional and geographical contexts (Du & Zhao, 2022; Q. Wang et al., 2022; X. Wang et al., 2018). For example, In Europe, where healthcare systems are predominantly public and geographically regulated, spatial accessibility and service quality emerge as primary drivers of hospital choice. In the Netherlands, patients with vascular diseases prioritize hospitals known for their reputable cardiology departments and low readmission rates for heart failure (Varkevisser et al., 2012). Dutch breast cancer and cataract patients show a preference for shorter travel distances even when hospital quality is slightly compromised (Salampessy et al., 2022). In the United Kingdom, an analysis of over 200,000 inpatient admissions across six hospitals in Derbyshire, revealed that distance was a stronger determinant of hospital choice than hospital size, waiting time, cleanliness, or facility availability (Smith et al., 2018). In China, where the healthcare system combines public provision with significant institutional hierarchy, insurance coverage and hospital level strongly shape patient preferences. A stratified random survey of 1,612 respondents aged 18 and above in Shenzhen, China, found that patients ranked their preferences as follows: insurance reimbursement, hospital level, distance, cost, and waiting time (X. Zhao et al., 2022). The United States presents a fundamentally different institutional landscape characterized by fragmented insurance systems, market-driven service delivery, and pronounced socioeconomic disparities. Orringer et al. (2022) found that race, insurance type, and household income were significantly associated with travel distance for elective total hip arthroplasty, with patients holding commercial or Medicare insurance and those from higher-income areas traveling farther to access care than Medicaid beneficiaries and lower-income patients.

Beyond hospital attributes, demographic and socioeconomic characteristics, corresponding to the predisposing dimension of Andersen's model, further shape healthcare-seeking behavior. A global survey spanning 16 countries, 14 medical specialties, and 32,651 participants found that individuals with higher SES tend to prioritize service quality and provider attitude over logistical considerations such as distance or cost. Conversely, individuals with lower SES or those from rural areas often emphasize logistical factors due to financial and accessibility constraints (Lin et al., 2024). Cultural and institutional factors also influence hospital preferences. In China, household registration (hukou) status significantly affects preferences for public versus private healthcare, with urban residents showing a stronger preference for public hospitals, while rural migrants display more varied choices (Tang et al., 2016). Other studies have also highlighted the impact of race, age, gender, and education level on healthcare preferences, underscoring the multifaceted nature of patient decision-making (Mhlanga & Hassan, 2022; Ohlson, 2020; W. Yu et al., 2017).

Despite these extensive findings, most prior research has examined individual or institutional factors in isolation, often overlooking how behavioral, socioeconomic, and spatial dimensions jointly influence visitation patterns across multiple facilities. Moreover, empirical models grounded in Andersen's framework rarely integrate fine-grained spatial interaction data capable of capturing population-level healthcare mobility. A deeper understanding of how spatial accessibility, social context, and perceived hospital attributes interact to shape healthcare visitation remains critical for advancing equitable and spatially informed health system planning.

**2.2 Models for Mobility Flow Prediction**

Mobility flow prediction plays a pivotal role in urban planning, transportation systems, public health, and emergency response. The increasing availability of extensive human mobility data sets has facilitated the emergence of data-driven predictive models. Traditional approaches, such as the gravity model inspired by Newton's law of gravitation, estimate flows based on population size and distance in a power-law form that can be linearized for estimation (Zipf, 1946). While the model is nonlinear in its mathematical form, it assumes monotonic and globally constant effects, thereby failing to represent the behavioral nonlinearities and spatial heterogeneity observed in complex mobility systems. Consequently, these traditional formulations are limited in capturing the contextual variability and dynamic complexity inherent in real-world human mobility. Machine learning (ML) techniques, such as Gradient Boosting, address these limitations by iteratively constructing predictive models capable of handling heterogeneous data sets and non-linear interactions (Guo et al., 2025). Unlike classical gravity models, ML approaches can flexibly learn from empirical data without predefined functional forms, improving predictive accuracy in heterogeneous environments.

Deep learning approaches further advance mobility flow prediction by uncovering complex patterns within large-scale data sets. Among these, Deep Gravity (Simini et al., 2021) exemplifies this progress by embedding gravity-law principles into a neural architecture, enabling end-to-end learning of spatial interaction patterns. The framework preserves the interpretability of traditional gravity models while capturing complex nonlinear dependencies among geographic and socioeconomic features (Cabanas-Tirapu et al., 2025). Subsequent studies have extended the Deep Gravity framework to a variety of mobility flow prediction tasks. For example, Yang et al. (2023) applied it to estimate intercity heavy-truck freight flows by integrating transportation infrastructure and land-use features, achieving substantial accuracy gains over traditional gravity and ML models. Zhao et al. (2025) adapted the model to simulate regional population movements by incorporating socioeconomic and spatial interaction factors, demonstrating its versatility in capturing both freight and human mobility dynamics. With explainable AI techniques, specifically SHAP as used in Deep Gravity, prior studies provide interpretable insights into model behavior, revealing the relative importance of geographic, infrastructural, and socioeconomic factors (Simini et al., 2021).

Graph Neural Networks (GNNs) provide a sophisticated methodology for mobility flow prediction by leveraging the inherent graph structure of spatial and temporal data. For instance, Wang et al. (2024) proposed a predictive model that combines GNNs to capture spatial dependencies with Long Short-Term Memory (LSTM) networks to model temporal trends. By incorporating travel mode choices and network analysis, this model predicts dynamic mobility flows with greater accuracy, effectively capturing both spatial and temporal patterns. Building on this foundation, Heterogeneous Graph Neural Networks (HGNNs) represent a cutting-edge approach to traffic flow prediction, capable of handling diverse, multi-modal data. An HGNN model integrates adaptive graph attention mechanisms and auxiliary "virtual" links to better capture spatial traffic patterns (T. Liu & Meidani, 2024). This model incorporates flow conservation principles into its loss function, ensuring accurate and consistent predictions. Experimental results highlight HGNNs' superior performance in accuracy, convergence, and generalization across different network topologies, positioning them as a robust solution for analyzing complex traffic flows (X. Liu et al., 2025; Xie et al., 2024). By combining adaptive attention with multi-modal feature fusion, HGNNs can jointly model spatial, functional, and semantic relationships within mobility networks.

Despite these methodological advances, the application of GNN- or HGNN-based models to healthcare system visitation flows remains underexplored, as most existing models predominantly address general traffic or commuting behavior (T. Liu & Meidani, 2024; B. Yu et al., 2018). Healthcare visitation, however, is unique in that it involves distinct spatial interaction patterns shaped by institutional attributes and patient demand characteristics, thereby requiring more specialized modeling approaches. Furthermore, the relative effectiveness of Deep Gravity models, an influential and widely adopted framework, versus HGNNs, an emerging paradigm in graph-based mobility modeling, in predicting healthcare visitation flows remains an open question, underscoring the need for comparative evaluations across model architectures within this domain.

## 3. Data and Method

## 3.1 Data Collection and Preprocessing

This study focuses on healthcare visitation within Harris County, Texas, which encompasses the City of Houston and surrounding suburban areas. All analyses and visualizations are conducted within Harris County, as healthcare facilities and patient catchment areas extend beyond the administrative boundaries of the city.

### 3.1.1 Healthcare Capacity and Occupancy Data

Hospital bed capacity data for Houston were obtained from CovidCareMap[1], a dashboard that integrates information from the Healthcare Cost Report Information System (HCRIS) to monitor healthcare system resources. The dataset includes 7,154 hospitals across the United States, spanning various types, such as General Acute Care Hospitals, Children's Hospitals, and Chronic Disease Hospitals. For this study, only General Acute Care Hospitals were considered, focusing on a subset of 50 facilities located in Houston. These hospitals represent the majority of healthcare utilization and reflect general population visitation behaviors. Other hospital types (e.g., specialty, rehabilitation, or psychiatric facilities) exhibit distinct service scopes and patient populations, leading to substantially different visitation patterns that should not be aggregated with general acute care services. The data set provides five key metrics critical for evaluating hospital resource availability and utilization: Staffed All Beds, representing the number of beds of all types routinely set up and ready for inpatient care; Staffed ICU Beds, which specifically refer to beds allocated for intensive care; Licensed All Beds, indicating the total number of beds approved for use by regulatory authorities; All Bed Occupancy Rate, reflecting the average percentage of hospital beds occupied; and ICU Bed Occupancy Rate, which shows the average percentage of intensive care unit beds in use. The spatial distribution of three bed capacity metrics in the Harris county (where Houston is located) is illustrated in Figure 1 (a), while two occupancy indices are shown in Figure 1 (b). Staffed All Beds ranged from 0 to 1,310, with a mean of 281.43; Staffed ICU Beds ranged from 0 to 162, with a mean of 30.23; and Licensed All Beds ranged from 4 to 1,403, with a mean of 368.97. For occupancy measures, the All Bed Occupancy Rate ranged from 0% to 86%, with a mean of 51.26%, while the ICU Bed Occupancy Rate ranged from 0% to 92%, with a mean of 43.24%.

---

[1] https://www.covidcaremap.org/maps/us-healthcare-system-capacity/#3.5/38/-96

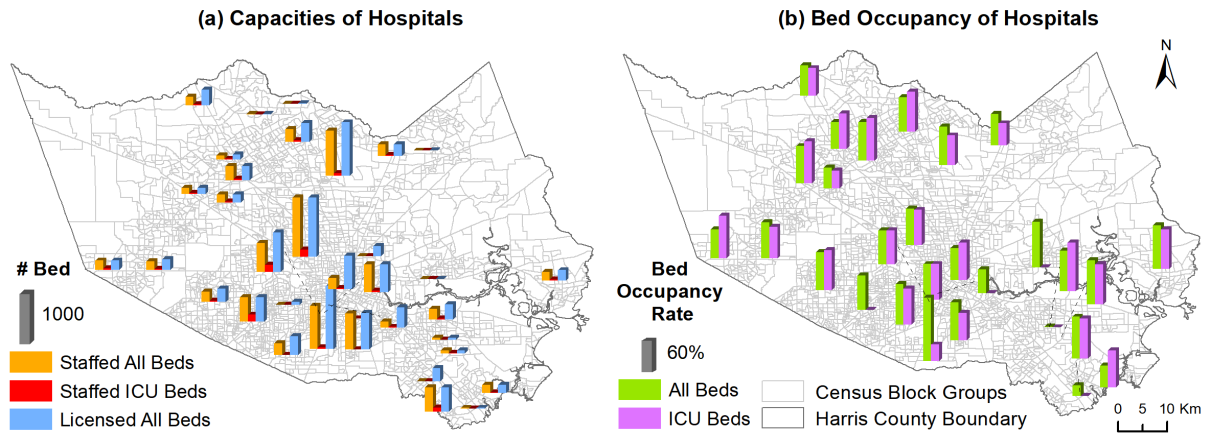

Figure 1 Spatial distribution of hospital capacities (a) and bed occupancy rates (b) (Gray bars serve as unified visual references for comparability across variables).

*3.1.2 Human Mobility Data*

The human mobility data were obtained from the Advan Foot Traffic / Weekly Patterns dataset (formerly SafeGraph) through the DEWEY data platform (Advan Research, 2025). This dataset aggregates anonymized origin-destination (OD) visitation data, capturing flows between residential Census Block Groups (CBGs) as origins and Points of Interest (POIs, including healthcare facilities) as destinations. Among the hospitals identified in the dataset, 35 General Acute Care Hospitals located in Harris County were matched with bed-capacity information from CovidCareMap and selected for analysis. Across these hospitals, a total of 1,657,488 individual visitations were recorded during the four-year period.

To construct an aggregated spatial interaction dataset, all weekly visitation records from January 2020 to December 2023 were aggregated from the SafeGraph dataset to the annual level, and then averaged across the four-year period. This produced a stable long-term representation of healthcare visitation intensity between residential areas and hospitals. Specifically, we summarized visitations from all 2,830 Census Block Groups (CBGs) in Harris County to the 35 hospitals included in the analysis, calculating the four-year average visitation volume for each CBG–hospital pair. The resulting dataset contains 16,783 distinct flows, as illustrated in Figure 2. Flow volumes ranged from 4 to 2,774.75, with an average of 24.69. High-volume flows were concentrated around eight hospitals, including three in the western region, one in the central area, and four in the eastern region. Most high-volume flows originated from areas near these hospitals, such as those surrounding the three hospitals in the southeastern corner.

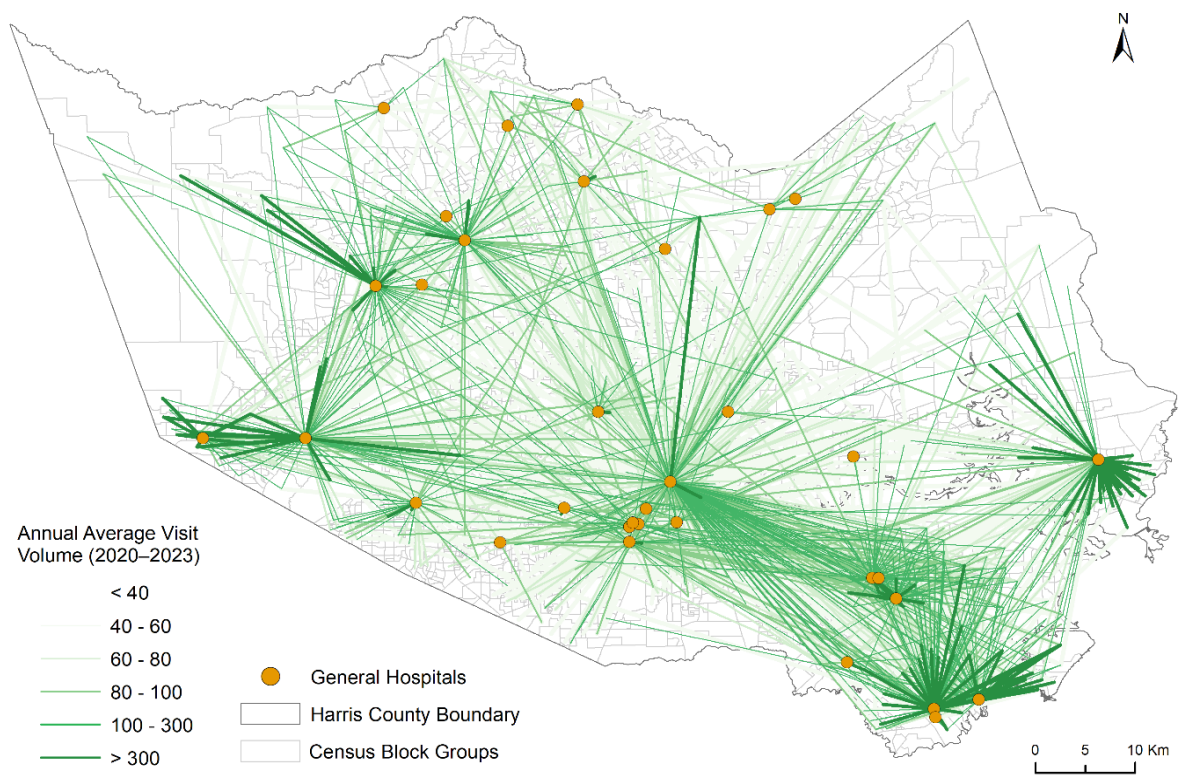

Figure 2. Spatial patterns of annual average visits to general hospitals in Houston (2020–2023)

However, there were notable exceptions, including long-distance flows, such as a four-year average visitation volume of 512.25 originating from Census Block Group FIPS 482019801001 (George Bush Intercontinental Airport) to St. Joseph Medical Center. These flows likely represent visits from residents outside Houston (a phenomenon known as Medical Tourism). Due to the unavailability of SES data for these flows, all flows originating from FIPS 482019801001 were excluded from the analysis, reducing the dataset to 16,730 flows.

In subsequent modeling, the SafeGraph visitation data were used as the dependent variable to approximate observed healthcare visitation behavior. However, as these data are aggregated from mobile device samples, they are inherently subject to sampling bias, uneven device coverage, and representativeness limitations, and therefore cannot be treated as ground-truth patient volumes. To mitigate these potential biases and ensure comparability across spatial units, we normalized the visitation data within each CBG by computing the proportion of visits from that CBG to each hospital relative to its total outgoing visits. This normalization emphasizes relative visitation preferences and spatial interaction patterns, allowing the modeling framework to capture the comparative structure of healthcare demand rather than the absolute magnitude of visits.

*3.1.3 Google Maps Data*

Two types of data were collected from Google Maps in 2024 for this study, reflecting the most recent publicly available information at the time of data acquisition. The first dataset pertains to attributes of 35 hospitals in Harris County, including their ratings and the number of review comments. These data were gathered using a web crawler and were not restricted to a specific historical period, as the healthcare seekers were used to capture each hospital's contemporary reputation and service perception rather than time-varying changes. Hospital ratings, displayed below the hospital name on the location information page (Figure 3), represent the average of all reviews. Ratings ranged from 1 to 4.8, with an average of 3.45. The number of review comments varied widely, ranging from 2 to 3,763, with an average of 966.74. The second dataset includes drive times from all CBGs to the 35 hospitals, obtained using the Distance Matrix API provided by Google Maps. Drive times ranged from 1.65 minutes to 69.95 minutes, with an average of 27.37 minutes.

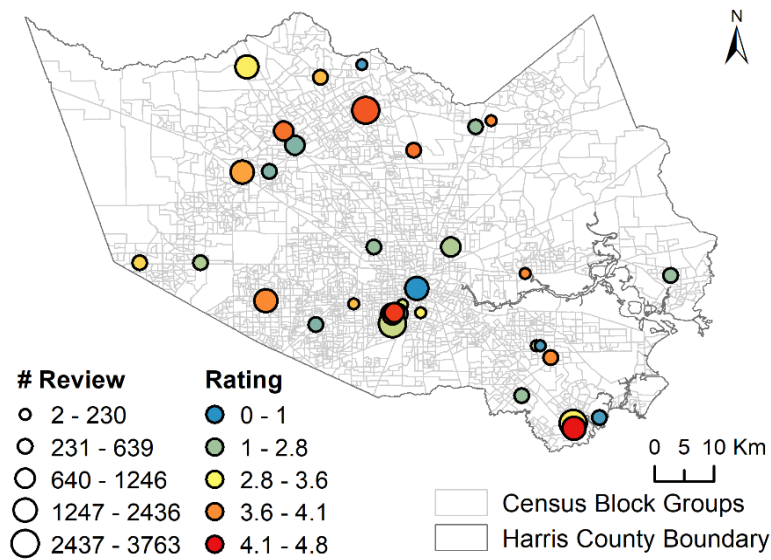

Figure 2 Spatial distribution of hospital user experience, where the number of reviews reflects hospital popularity and ratings represent perceived service quality

It is important to note that Google Maps reviews may reflect demographic and behavioral biases, as younger and more digitally active individuals are more likely to contribute feedback. However, these data remain valuable for this study because online ratings and review counts directly influence how people perceive and choose healthcare facilities. In this sense, they capture the information environment that shapes healthcare-seeking behavior rather than objective service quality. Therefore, ratings and reviews are interpreted as proxies for social perception and public awareness within the healthcare system. Moreover, the number of reviews itself conveys behavioral meaning, reflecting each hospital's visibility and perceived credibility. Patients often regard hospitals with more reviews as more trustworthy when ratings are comparable. Hence, variation in review volume represents a meaningful perceptual signal rather than a data limitation.

*3.1.4 Socioeconomic Data*

We collected socioeconomic data for 2,830 census block groups in Harris County from the U.S. Census American Community Survey 5-Year Data (2009–2022). The variables were selected based on well-established determinants of healthcare accessibility and utilization identified in prior studies, including age composition, racial and ethnic structure, educational attainment, income, and vehicle ownership, which are widely recognized as key predictors of healthcare-seeking behavior and mobility patterns (Guagliardo, 2004; Luo & Wang, 2003; F. Wang & Luo, 2005). The final set of indicators included total population, percentage of the population under 18, percentage over 65, percentages of Hispanic, White, Black, and Asian populations, percentage with a bachelor's degree or higher, median household income, and percentage of households with a vehicle.

Table 1. Summary of Variables, Categories, and Data Sources

| Categories | Variables | Data Sources |
|---|---|---|
| Healthcare Systems Attributes | Staffed All Beds | CovidCareMap Dashboard |
| | Staffed ICU Beds | |
| | Licensed All Beds | |
| | All Bed Occupancy Rate | |
| | ICU Bed Occupancy Rate | |
| | # Reviews | Google Maps Review Data |
| | Rating | |
| | Hospital Longitude | |

|  | Hospital Latitude | CovidCareMap Dashboard |
|---|---|---|
| Population Socioeconomic Status | Total Population | American Community Survey |
|  | % Population Under 18 |  |
|  | % Population Over 65 |  |
|  | % Hispanic Population |  |
|  | % White Population |  |
|  | % Black Population |  |
|  | % Asian Population |  |
|  | % Bachelor's degree or Higher |  |
|  | Median Household Income |  |
|  | % Households with a Vehicle |  |
|  | Visitor Departure Longitude | US Census |
|  | Visitor Departure Latitude |  |
| Distance Cost | Drive Time | Google Map Data |
| Hospitals Visitation | Visitation Percentage | SafeGraph Data |

In summary, as shown in Table 1, the modeling includes 9 attributes of healthcare systems and 12 SES attributes of the population at the census block group level. The dependent variable is the Visitation Percentage. Drive time is used as the metric for distance cost between block groups and hospitals. In addition, the geographic coordinates (longitude and latitude) of hospitals and census block groups were included to capture broader spatial structures not represented by drive time alone. While drive time measures direct travel distance, spatial relationships among hospitals and among residential areas, such as competition and clustering, also influence visitation patterns. Incorporating coordinates enables the model to learn these implicit spatial dependencies without explicitly constructing distance matrices. This practice is common in deep spatial modeling, where longitude and latitude act as positional encodings to preserve spatial context (Simini et al., 2021; Song, 2023).

## 3.2 Predictive Models for Healthcare Facility Visits

*3.2.1 Model Descriptions and Experimental Settings*

Various models, including Gradient Boosting, Multi-Layer Perceptrons (MLPs), Deep Gravity, and Heterogeneous Graph Neural Networks (HGNN), were employed for flow prediction tasks. To establish a benchmark for comparison, a Naive Regression model was included as a baseline. For the Naive Regression and Gradient Boosting models, visitation percentage served as the dependent variable, while the independent variables consisted of nine hospital attributes, 12 SES attributes of the population, and one drive time attribute. The MLP model, a fully connected neural network, was designed to capture non-linear relationships between input features and the target variable. This model processed input features, including block group attributes, hospital attributes, and drive time, through multiple hidden layers. Using backpropagation, it optimized weights to minimize the loss function.

The Deep Gravity model extended the traditional gravity model by integrating neural network techniques to account for complex, non-linear interactions among block group attributes, hospital attributes, and drive time. In this framework, the attributes of block groups, hospitals, and drive time were first encoded using MLPs. The encoded features were then concatenated and passed through another MLP for decoding and final prediction, as shown in Figure 4. Hyperparameter tuning was performed to optimize the model structure. After testing various configurations, including different embedding strategies, the current structure demonstrated the best performance.

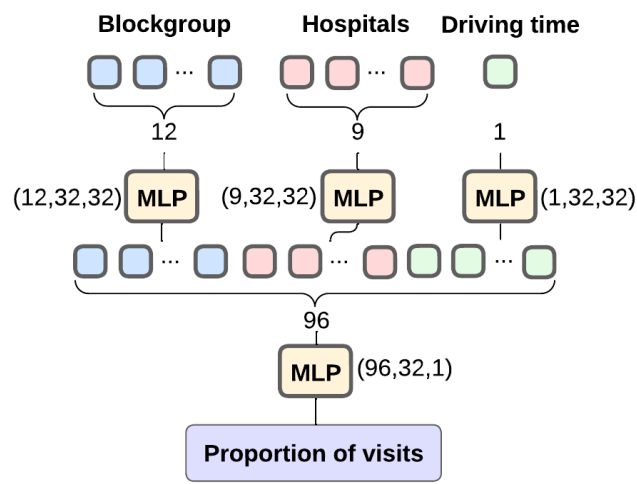

Figure 4. Structure of the Deep Gravity Model

The HGNN model employed a graph-based approach to incorporate both node-level and edge-level attributes in the prediction process. In this model, hospitals and block groups were represented as two distinct types of nodes, with their respective attributes serving as node features. Drive time between block groups and hospitals was modeled as an edge attribute (Figure 5(a)). The model processed block group and hospital attributes, along with drive time encoding, through MLPs to generate initial embeddings. These embeddings were passed through a heterogeneous SAGEConv network to extract features for block groups and hospitals. The final features were combined using weights for block group attributes, hospital attributes, and drive time (denoted as weight_B, weight_H, and weight_D, respectively). Through tuning, the weight combination of 1, 1, and 4 delivered the best results. Additionally, experiments with alternative structures, such as Graph Attention Networks for attribute embedding, confirmed the superiority of the current configuration.

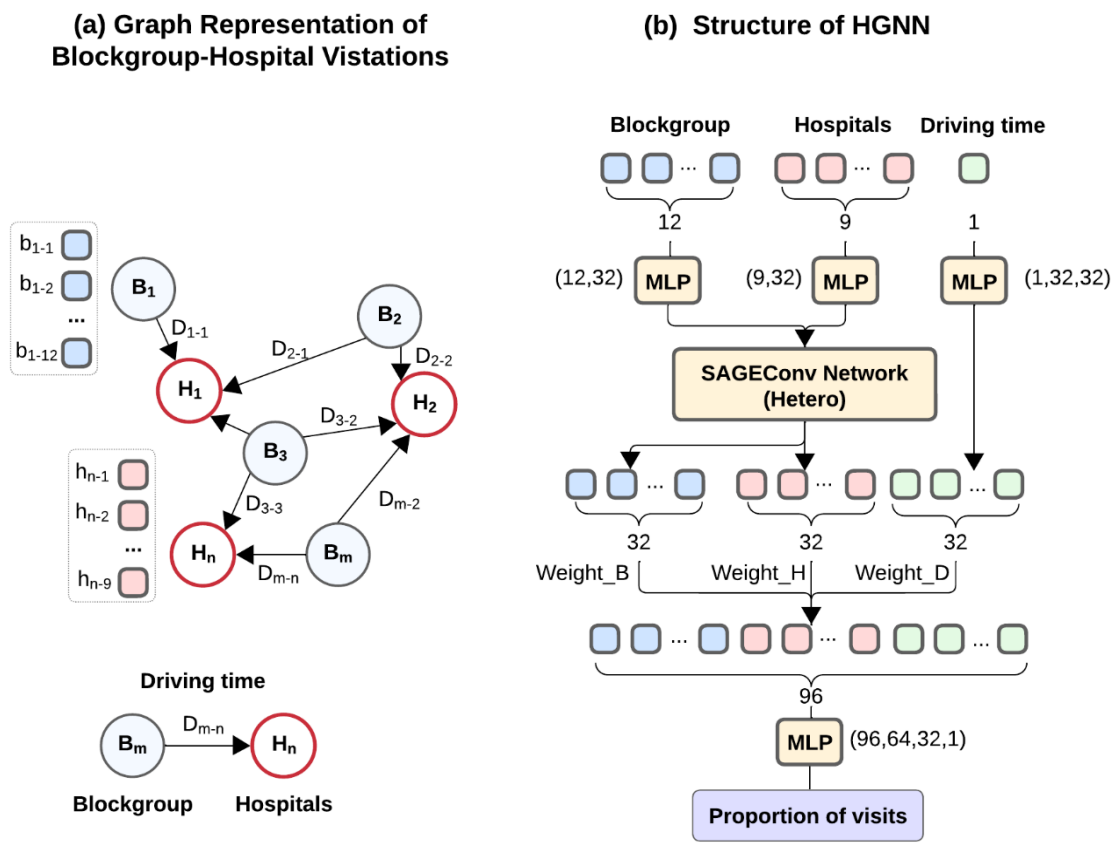

Figure 5. Graph Representation and Structure of HGNN Model

For all models, 90% of the data was allocated for training and 10% for testing. To ensure robust and unbiased evaluation, each model was trained and validated using 10 independent runs under a 10-fold cross-validation scheme, and the averaged training and validation losses across runs were recorded. Figure 6 illustrates the convergence patterns of the MLP, HGNN, and Deep Gravity models, where the light lines represent individual runs and the bold lines indicate the mean losses. To maintain a balance between training and validation performance, the MLP model was trained for 800 epochs, while the Deep Gravity and HGNN models were each trained for 400 epochs.

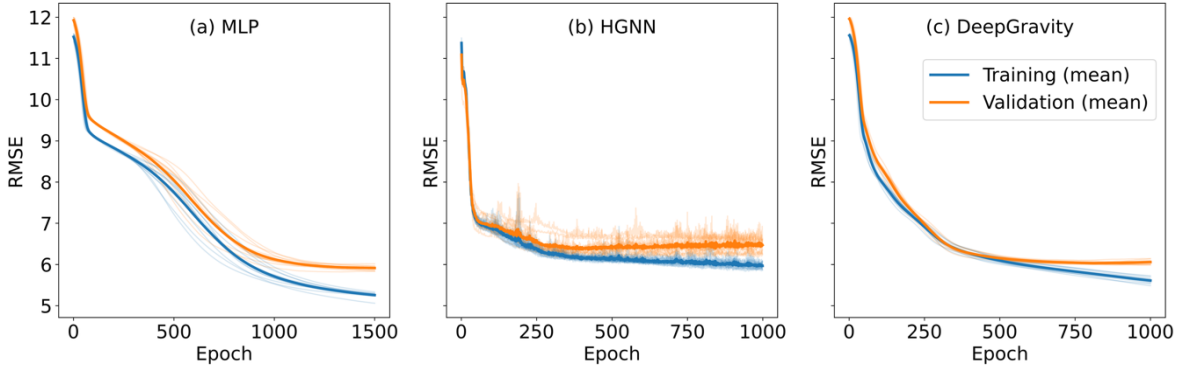

Figure 6. Training and Validation Loss over Epochs for MLP, HGNN, and Deep Gravity Models (light lines = individual runs; bold lines = mean performance across 10 runs).

### 3.2.2 Performance Evaluation Metrics

We evaluated the performance of each model on the same test set using three widely recognized metrics for mobility flow prediction: Normalized Root Mean Square Error (NRMSE) (Equation 1), Symmetric Mean Absolute Percentage Error (SMAPE) (Equation 2), and Common Part of Commuters (CPC) (Equation 3) (Zhang et al., 2024). NRMSE measures the average deviation between predicted and actual values, normalized by the range of the data, with a value range of [0, ∞), where lower values indicate better accuracy. SMAPE calculates the symmetric percentage error between predicted and actual values, ensuring unbiased evaluation, with a range of [0,200%], where smaller percentages represent better performance. CPC assesses the overlap between predicted and actual commuter flows, focusing on distributional similarity, with values ranging from 0 to 1, where higher values signify greater similarity.

$$NRMSE = \frac{\sqrt{\frac{1}{N}\sum_{i=1}^{N}(y_i - \hat{y}_i)^2}}{\max(y) - \min(y)} \quad (1)$$

$$SMAPE = \frac{1}{N}\sum_{i=1}^{N}\frac{|y_i - \hat{y}_i|}{(|\hat{y}_i| + |y_i|)/2} \times 100 \quad (2)$$

$$CPC = \frac{2\sum_{i=1}^{N} \min(y_i, \hat{y}_i)}{\sum_{i=1}^{N}(y_i + \hat{y}_i)} \tag{3}$$

Where $\hat{y}_i$ is the predicted value for flow $i$, $y_i$ is the actual value for flow $i$, N is the total number of flows, $\max(y)$ and $\min(y)$ are the maximum and minimum values of the actual data, respectively. To ensure a robust and unbiased evaluation, all models were trained and validated using 10-fold cross-validation, and the average performance metrics across all folds are reported in Section 4.1.

### 3.3 Methods of Analysis

To analyze the factors influencing visitation patterns, we employed SHAP (SHapley Additive exPlanations) for model interpretation and a Partial Dependence Plot (PDP) (Friedman, 2001) approach for analysis of specific factors or their interactions. SHAP is a widely used, model-agnostic technique rooted in cooperative game theory, where each feature in a predictive model is assigned a "Shapley value" that represents its contribution to the model's output (Simini et al., 2021). It systematically evaluates all possible combinations of factors, measuring how the prediction changes when a specific factor is added or removed from different subsets. This comprehensive approach ensures that each factor's influence is assessed across all scenarios, capturing potential interactions with other variables. By averaging these changes, SHAP provides a fair and consistent attribution of factor importance, reflecting both the magnitude (the strength of the factor's impact) and the direction (whether it increases or decreases the predicted value). In the context of visitation patterns, Shapley values reveal how hospital characteristics and block group attributes collectively shape the model's predictions. In our analysis, we further applied group SHAP to address multicollinearity among highly correlated hospital capacity variables. Specifically, staffed all beds and licensed all beds exhibited strong correlation and separating them could lead to unstable or misleading attributions. To ensure more robust interpretation, these variables were combined into a single group named "All beds", and the resulting Shapley value represents the aggregate contribution of this capacity-related factor. This grouped representation provides a clearer and more reliable assessment of how overall hospital bed availability influences visitation patterns.

We employed the Partial PDP method to conduct scenario-based hypothetical simulations, each assuming a specific hospital and block group configuration to examine how visitation patterns would emerge based on human mobility dynamics learned by the optimal model. PDP is a widely used tool for interpreting black-box predictive models, as it isolates the effect of a single variable on the outcome while keeping all other factors constant (Apley & Zhu, 2020; Inglis et al., 2022; Nohara et al., 2022). By eliminating confounding influences, PDP facilitates a clearer understanding of the individual impact of a predictor on the model's output under controlled assumptions. To analyze how visitation patterns respond to variations in hospital or block group attributes, we conducted multiple PDP-based simulations, each modifying a single attribute while keeping all others fixed at their average values. Consistent with our SHAP analysis, the hospital capacity measure used in these simulations was the grouped "All beds" variable, which combines staffed all beds and licensed all beds due to their high correlation. For example, when examining the effect of hospital rating, we assigned the rating to maximum, average, and minimum values while holding all other attributes, such as hospital capacity and block group demographics, constant at their means. Meanwhile, to model distance decay, we varied drive time from 0 to 70 minutes while keeping all other attributes constant, observing how visitation percentages declined with increasing travel distance. This approach allowed us to generate three distinct distance decay curves, each representing a different hypothetical scenario where only the hospital rating varied while all other factors remained unchanged.

A key insight from these simulations is the identification of an inflection point, where the distance decay curves corresponding to different hospital attributes scenarios intersect. This inflection point represents a balance between hospital attributes and spatial accessibility, illustrating how variations in hospital attributes interact with distance-based decision-making. It is important to note that this inflection point does not indicate a sharp threshold. Instead, it reflects a gradual transition in decision-making priorities revealed by the trained models. The precise location of this balance point is not fixed; it may vary depending on regional healthcare availability, transportation infrastructure, and patient demographics.

## 4. Results

### 4.1 Model Performance

Table 2 reports the mean and standard deviation of NRMSE, SMAPE, and CPC across 10-fold cross-validation, reflecting both model accuracy and stability. The Deep Gravity model outperformed all other models, achieving the lowest mean NRMSE (0.62±0.0036) and SMAPE (59.71±1.0256), as well as the highest CPC (0.74±0.0059), demonstrating its relative superior accuracy and effectiveness in predicting the healthcare system visitation flow. The Naive Regression model exhibited the weakest performance across all metrics, with an average

NRMSE of 1.00, SMAPE of 86.32, and CPC of 0.58. Gradient Boosting displayed moderate results, with an average NRMSE of 0.80, SMAPE of 70.21, and CPC of 0.67. The MLP and HGNN models achieved identical average NRMSE scores (0.68), with MLP showing slightly better results in SMAPE (70.52 compared to 79.64) and CPC (0.70 compared to 0.69). Although HGNN is the most advanced and complex model among those models, its performance did not surpass that of the Deep Gravity model. This finding underscores that more advanced models are not necessarily superior; instead, selecting a model that aligns with the specific characteristics and requirements of the task is crucial for achieving optimal results. Overall, the Deep Gravity model demonstrated the best performance and the greatest stability, and the subsequent interpretation and analysis are based on its results.

Table 2. Average Model Performance (mean ± standard deviation) Based on 10-Fold Cross-Validation[2]

| Models | NRMSE | SMAPE | CPC |
| --- | --- | --- | --- |
| Naive Regression | 1.00±0.0002 | 86.32±0.9717 | 0.58±0.0332 |
| Gradient Boosting | 0.80±0.0128 | 70.21±0.4663 | 0.67±0.0040 |
| MLP | 0.68±0.0247 | 70.52±1.5807 | 0.70±0.0101 |
| **Deep Gravity** | **0.62**±0.0036 | **59.71**±1.0256 | **0.74**±0.0019 |
| HGNN | 0.68±0.0167 | 79.64±2.1719 | 0.69±0.0059 |

Figure 7 illustrates the spatial distribution of observed and predicted flows generated by the Deep Gravity model, based on the test dataset. The observed flow's visit percentage ranged from 0.10% to 83.97%, while the predicted flow's visit percentage ranged from 0% to 69.39%. The average absolute error between the observed and predicted flows was 3.85%, reflecting the model's strong predictive accuracy in capturing healthcare system visitation patterns. The predicted flows closely align with the observed flows in the spatial pattern. Hospitals located near the city center exhibited higher visit percentages, while hospitals in peripheral areas displayed lower visit percentages, matching the observed data. Furthermore, the model successfully captured the overall visitation flows pattern of several hospitals outside the central urban area, demonstrating its ability to account for moderate visitation flows in peripheral locations.

---

[2] Values are reported as mean ± standard deviation across 10-fold cross-validation. Lower values of NRMSE and SMAPE indicate better performance, while higher CPC values indicate better performance.

The model displayed a tendency to underestimate visit percentages in longer distance flows. For instance, St. Joseph Medical Center in the central region had 21 flows with observed visit percentages exceeding 40%, which were underestimated by an average of 27.40%. Houston Methodist San Jacinto Hospital in the easternmost region had 13 flows with observed visit percentages exceeding 40%, which were underestimated by an average of 19.35%.

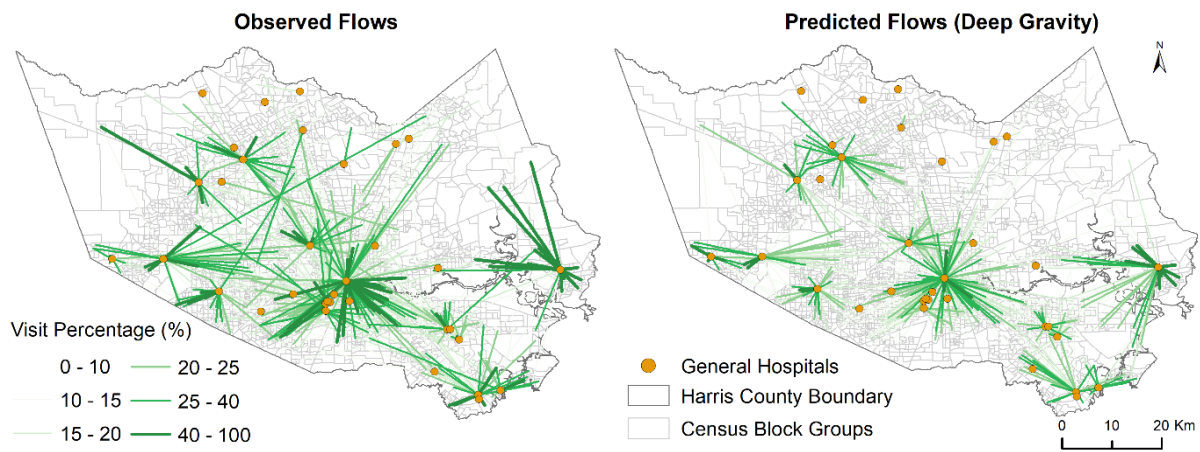

Figure 7. Observed vs. Predicted Flows by the Deep Gravity Model on the Test Dataset

## 4.2 Factors Influencing Hospital Visitations

Figure 8 provides a visual summary of the top 20 factors influencing hospital visitation percentages, ranked by their importance based on SHAP values derived from the Deep Gravity Model. Each bar's length represents the magnitude of influence, with longer bars indicating stronger effects on visitation percentages. The red-to-blue gradient within each bar conveys the actual values of the corresponding variable, where red represents higher values and blue represents lower values. The SHAP value itself reflects the direction and strength of each factor's impact on model output. A positive SHAP value indicates the positive relationship between the variable and model output, and vice versa. As shown in Figure 7, the most influential factor is driving time, showing a strong negative relationship with visitation percentages. This is consistent with the distance decay pattern revealed in the previous hospital flow prediction studies (Jia et al., 2019; F. Wang, 2021), where longer travel times significantly decrease the likelihood of hospital visits. Spatial factors, including hospital latitude and longitude and the Visitor departure latitude and longitude, further emphasize the role of geography in healthcare visitation spatial patterns.

Beyond distance, hospital-related attributes exhibit notable associations with visitation patterns. ICU Bed Occupancy Rate is the second most influential factor after driving time, showing a positive relationship with visitation percentages. All beds, ranked third, display nonlinearly associations with visitation percentages. It is essential to acknowledge that these observed relationships represent overall univariate correlations and may be substantially influenced by other interacting factors and contextual conditions. The combined effects of hospital attributes and distance, along with a detailed interpretation of these relationships, are further examined in Section 4.3. Concurrently, user experience factors play a significant role in shaping visitation patterns. The number of review comments and hospital ratings rank fourth and sixth, respectively. The number of review comments generally exhibit a positive association with visitation, indicating that hospitals receiving more feedback tend to attract greater patient volumes. Alternatively, this relationship may indicate that hospitals with greater visitation naturally accumulate more comments due to a larger patient base. In contrast, hospital ratings reveal a more nuanced pattern, showing an overall negative correlation with visitation percentages. This negative relationship may shift when additional factors, such as distance and population characteristics, are considered, underscoring the importance of a comprehensive, multifactorial analysis.

Population characteristics also play a significant role in shaping visitation patterns, supporting our hypothesis that visit patterns are influenced not only by distance decay but also by hospital attributes and residents' SES. Factors such as the percentage of Hispanic population, percentage of Black population, and total population rank prominently in Figure 8. The percentage of the Black or Hispanic population shows an overall positive correlation with visitation percentages, while the total population do not exhibit a clear positive or negative trend.

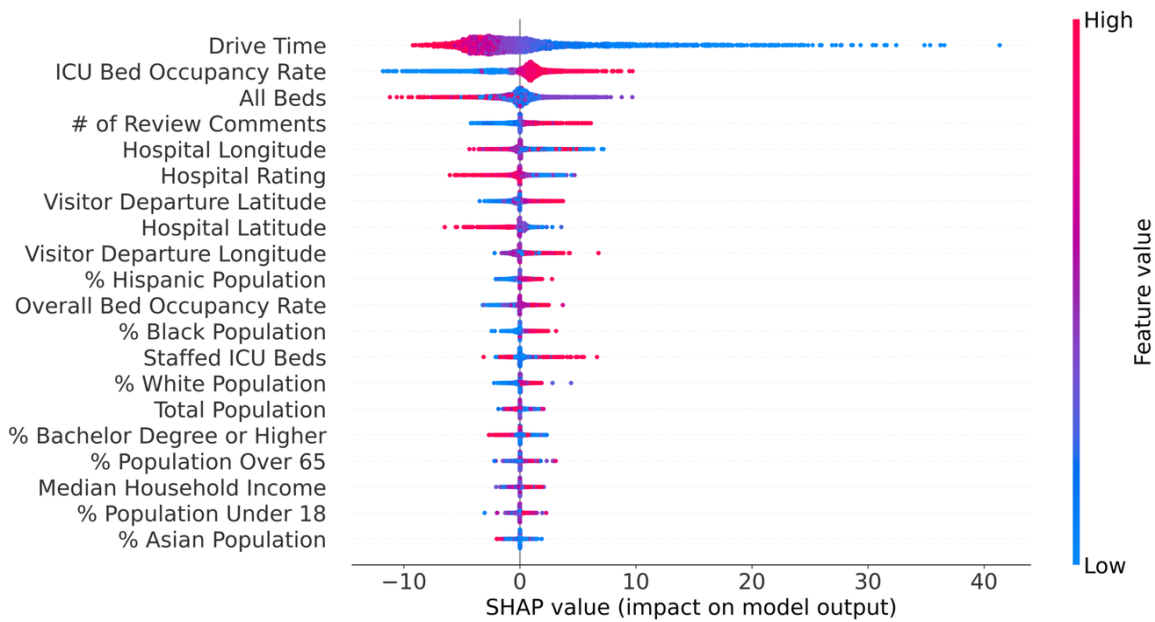

Figure 8. Top 20 Factors Influencing Hospital Visitation Shown by the SHAP Summary Plot of the Deep Gravity Model

### 4.3 Impacts of Hospital Attributes on Visitation Patterns

The analysis, as shown in Figure 9, illustrates the impact of hospital attributes on distance decay patterns. The blue line represents the simulated pattern when variables are set at their average values, while the yellow and red lines reflect patterns at their maximum and minimum values, respectively. All other hospital and block group attributes are fixed at their average values to isolate the effect of the varying hospital attributes on distance decay. A clear distance decay trend is evident across all scenarios, consistent with findings in Figure 8, which identified that driving time is significantly and negatively related to visitation proportions. Variations in hospital attributes reveal trade-offs that patients may weigh when choosing hospitals, with inflection points where the impact of these attributes shifts based on distance.

The All beds measure, constructed by combining staffed all beds and licensed all beds to mitigate multicollinearity, together with Staffed ICU Beds provides two representations of hospital size and resource availability. When examining their PDP results jointly, a consistent distance-dependent pattern emerges. At short travel distances, visitation probabilities tend to be higher for medium-sized or smaller hospitals, suggesting that proximity-driven decisions favor facilities that are easier to access or perceived as more convenient. For the All beds variable, this relationship shifts beyond an approximate inflection point at around 32 minutes, after which larger hospitals become relatively more attractive. This indicates that once patients are already traveling longer distances, they may be more inclined to seek hospitals with broader overall resources. The Staffed ICU Beds results show a similar general distance–preference pattern but without a clearly identifiable inflection point within the modeled range.

ICU bed occupancy exhibits a distance-dependent influence on hospital visitation patterns. Within a 38-minute travel time, higher ICU occupancy is associated with increased visitation. This trend may reflect two potential mechanisms: first, patients might perceive high ICU occupancy as an indicator of a hospital's expertise and capability in managing critical cases; second, a higher number of ICU patients could result in more friends or family members visiting them, thereby increasing overall visitation. Beyond this threshold, higher ICU occupancy correlates with lower visitation rates, suggesting that patients prioritize immediate access to available ICU resources when travel distances increase. Similarly, overall bed occupancy shows a positive correlation within 22 minutes, possibly reflecting a perception of reliability. However, beyond 22 minutes, this correlation turns negative, which suggests that patients might associate high occupancy with overcrowding. The number of reviews follows a similar trend, exhibiting a positive relationship with visitation percentage within 34 minutes, as patients tend to rely more on review counts as a key decision-making factor. However, beyond this threshold, the relationship reverses, with higher review numbers associated with lower visitation percentages. This shift suggests that as distance increases, the importance of review numbers diminishes, and other factors, such as resource availability, may become more influential in hospital choice.

Hospital ratings exhibit a distinct pattern. Low-rated hospitals experience a more pronounced distance decay effect, whereas highly rated hospitals show a weaker distance decay. This suggests that highly-rated hospitals have a stronger ability to attract long-distance patients compared to lowly rated hospitals. In terms of the inflection point, within approximately 19 minutes, ratings may have a limited influence. This phenomenon could be interpreted as patients prioritizing convenience over quality indicators or perceiving quality differences as less critical for shorter trips. Beyond this threshold, hospital ratings become increasingly significant, as higher ratings provide reassurance of quality. This is particularly relevant for long-distance patients, who may seek additional justification for extended travel time.

These findings indicate that while distance decay is a universal trend, hospital attributes shape its specific patterns. For shorter distances, convenience and stability may take precedence, with attributes such as high occupancy and numerous reviews potentially being interpreted as indicators of reliability. At longer distances, excellence assurance factors, such as ratings and perceived resource availability, might become more influential. These dynamic trade-offs offer insights that could inform strategies for optimizing healthcare resource allocation and policy development.

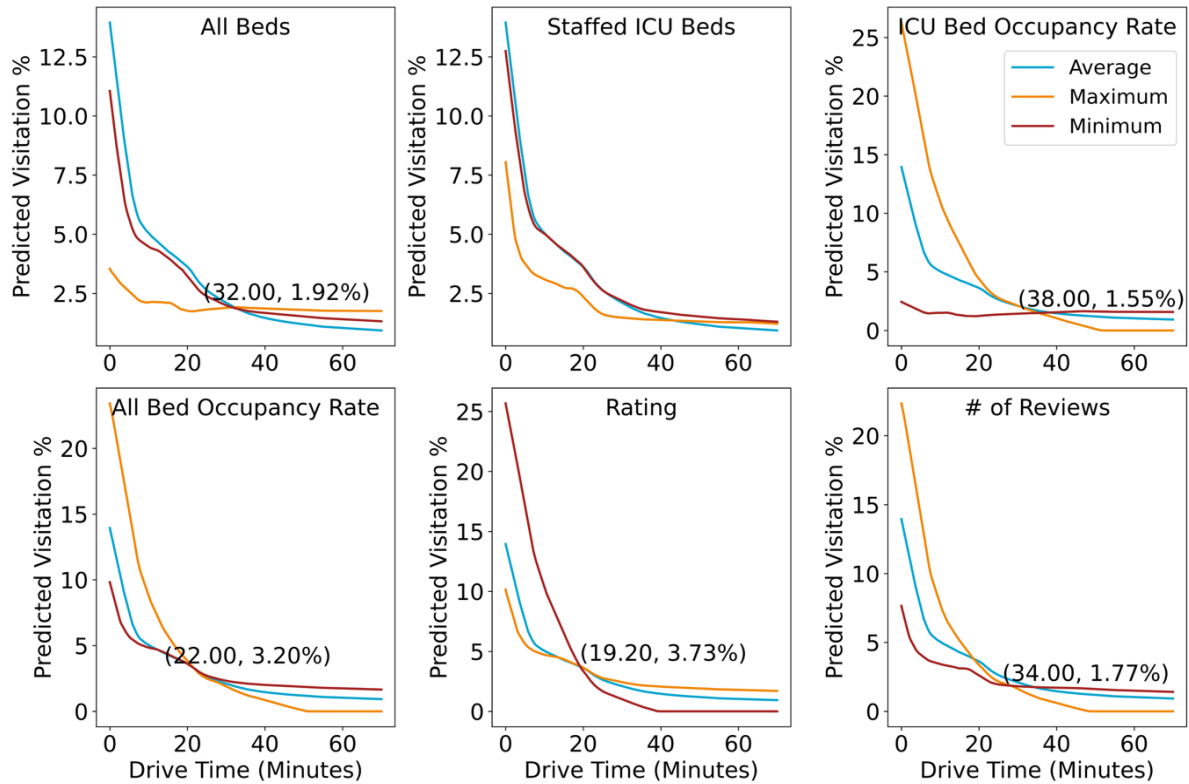

Figure 9. Effects of Hospital Attributes on Visitation Distance Decay Patterns

## 4.4 Combined Effects of Hospital Ratings and Population SES on Visitation Dynamics

Figure 10 illustrates the trade-off between hospital ratings and drive time across areas with varying population characteristics, highlighting the combined effects of hospital ratings and population SES on visitation distance decay patterns. Overall, regions with higher proportions of Hispanic, Black, under-18, and over-65 populations exhibit higher hospital visitation percentages compared to regions with lower representations of these demographic groups. These disparities may stem from the higher healthcare demands of these groups.

Across all groups, a critical inflection point emerges where the relationship between hospital ratings and visitation shifts from negative to positive. This threshold varies by demographic factors. In predominantly White areas, the inflection point occurs later at 32 minutes, whereas in non-White areas, it emerges much earlier at 18 minutes. This suggests that patients in predominantly White areas exhibit a higher tolerance for travel time before hospital ratings significantly influence their decisions. In contrast, patients in non-White areas may place greater weight on hospital ratings at shorter distances, leading to an earlier shift in visitation preferences. Asian-dominated areas exhibit the earliest inflection point (8 minutes) compared to non-Asian areas (19 minutes), highlighting the Asian population's significant sensitivity to hospital reputation, influencing their consideration of ratings even for very short travel distances. Educational attainment also shapes healthcare choices significantly. In regions with higher educational levels, the inflection point occurs at 12 minutes, compared to 21 minutes in areas with lower education levels. This suggests that highly educated populations prioritize hospital quality even within shorter travel distances, whereas less educated groups may rely more on geographic convenience and exhibit lower sensitivity to ratings.

Hispanic population, Black population, age, and income levels do not significantly influence the balance between hospital ratings and distance. Regions with predominantly Hispanic or Black populations exhibit slightly earlier inflection points, at 17 and 18 minutes, respectively, compared to areas with lower proportions of these populations, where the inflection points occur at 19 and 18 minutes, respectively. Block groups composed entirely of individuals under 18 or over 65 exhibit earlier inflection points, both at 15 minutes, compared to 18 and 17 minutes in areas without these age groups. High-income areas experience a slightly later inflection point (18 minutes) compared to low-income areas (17 minutes).

These findings highlight the intricate disparities in healthcare access resulting from the interaction between hospital ratings and population characteristics, particularly the proportions of White or Asian populations and education levels. Policymakers could consider the unique needs of different communities when designing and implementing strategies for healthcare resource allocation to promote equity in access and health outcomes.

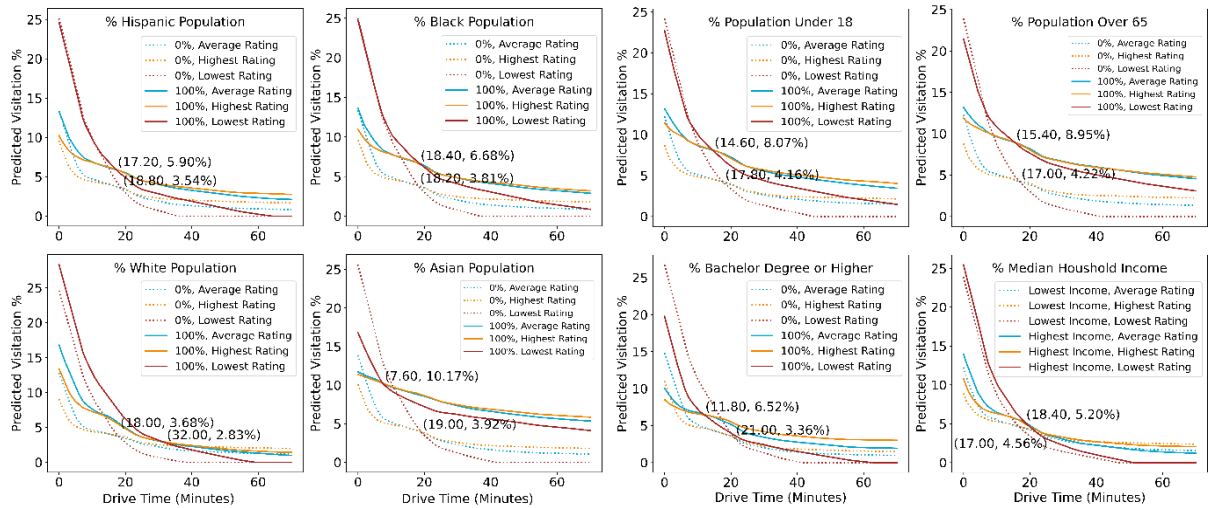

Figure 10. Combined Effects of Hospital Ratings and Population SES on Visitation Distance Decay Patterns

## 5. Discussion

### 5.1 Significant Implications

This study offers several significant implications for advancing our understanding of healthcare visitation patterns. First, this study incorporates a comprehensive range of factors, including hospital attributes—such as capacities, occupancy rates, reputation, popularity, and geographic location—alongside population characteristics like race, age, education, income, and geographic distribution to predict healthcare visitation flows. SHAP-based analysis highlights the significant influence of key factors, including driving time, hospital capacities, reputation, and popularity, on visitation flow distribution, many of which have been overlooked in previous studies. Notably, this study uncovers interactive effects between hospital ratings and population characteristics, providing new insights into visitation patterns that prior research has not thoroughly explored.

Second, the study leverages social sensing data to enhance flow prediction and healthcare accessibility analysis. By incorporating user-generated ratings and reviews from Google Maps, it captures patient-centered perspectives on hospital quality, enriching the characterization of hospital attributes. Additionally, the use of real human mobility data from SafeGraph, spanning four years, enables a detailed and granular examination of visitation patterns across communities. This robust dataset provides a foundation for predicting healthcare visitation flows with high precision and reliability.

Third, the study evaluates the performance of four widely used flow prediction models in healthcare system analysis: Gradient Boosting, MLPs, Deep Gravity, and HGNN, with a naive regression model serving as baseline. While HGNN is the most advanced and sophisticated model among those tested, its performance does not exceed that of the Deep Gravity model. This outcome can be partly attributed to the structural simplicity of the healthcare visitation network, which consists of only two node types, hospitals and CBGs. Deep Gravity, as a shallower architecture directly modeling pairwise feature interactions, is better suited for such bipartite networks, whereas HGNNs typically demonstrate superior performance in more complex, multi-relational graphs involving diverse node and edge types. This finding highlights the importance of aligning model selection with the specific characteristics and requirements of the research task rather than assuming that the most complex model will invariably yield superior results.

Fourth, the results validate the hypothesis that hospital visitation patterns are shaped by a combination of distance, hospital attributes, and population SES. Hospital capacity, ICU bed occupancy, popularity, and quality ratings are significant predictors, with their influence varying across travel distances—convenience and stability factors, such as high occupancy rates, are more critical for shorter distances, while indicators of excellence, such as high ratings, gain prominence for longer distances. SES further influence these patterns, as areas with higher proportions of Hispanic, Black, under-18, and over-65 populations tend to have more frequent hospital visits, potentially reflecting greater healthcare needs or limited access to alternative medical services. White-majority areas exhibit lower sensitivity to hospital ratings for shorter distances, whereas Asian populations demonstrate heightened sensitivity to hospital reputation, even for nearby facilities. Highly educated populations prioritize hospital quality at shorter distances, while less educated groups tend to emphasize geographic proximity and show reduced sensitivity to quality ratings.

Finally, these findings provide critical insights for policymakers to address healthcare access disparities. Enhancing hospital quality and reputation, particularly in areas serving populations such as Asians and highly educated groups, who exhibit a strong sensitivity to hospital ratings, can ensure more equitable access to healthcare resources. Improving geographic accessibility for communities with lower SES and less education by increasing the availability of local healthcare facilities, especially primary care services, can help address their reliance on proximity-based care. Allocating resources such as ICU beds, specialized services, and staff training to underserved areas with higher proportions of Hispanic, Black, under-18, and over-65 populations is also essential to address their specific healthcare needs. These strategies not only reduce healthcare access inequality but also contribute to building sustainable and inclusive healthcare systems that serve all populations effectively.

**5.2 Limitations and Future Research**

This study has several limitations that highlight opportunities for future research. One key limitation is the absence of insurance coverage as a factor in the analysis. Insurance coverage is widely recognized as one of the strongest predictors of healthcare accessibility and visitation behavior, yet detailed insurance data at the required spatial granularity were unavailable. Its exclusion likely contributes to the moderate performance of all models, as key behavioral and financial determinants remain unobserved. Incorporating such data in future research could substantially enhance model accuracy and yield a more comprehensive understanding of how financial barriers intersect with hospital characteristics and population socioeconomic status.

Second, the study relies on a four-year average of visitation data, which overlooks temporal variations and evolving relationships over time. This limitation is particularly relevant in the context of the COVID-19 pandemic, which has drastically altered healthcare utilization patterns (Figure 11). Compared to the pandemic period, the most notable difference during the recovery phase is the overall increase in healthcare visitation volume, particularly in long-distance healthcare visits. Future work should explore how visitation patterns and influencing factors have changed during and after the pandemic, providing insights into the dynamic nature of healthcare access.

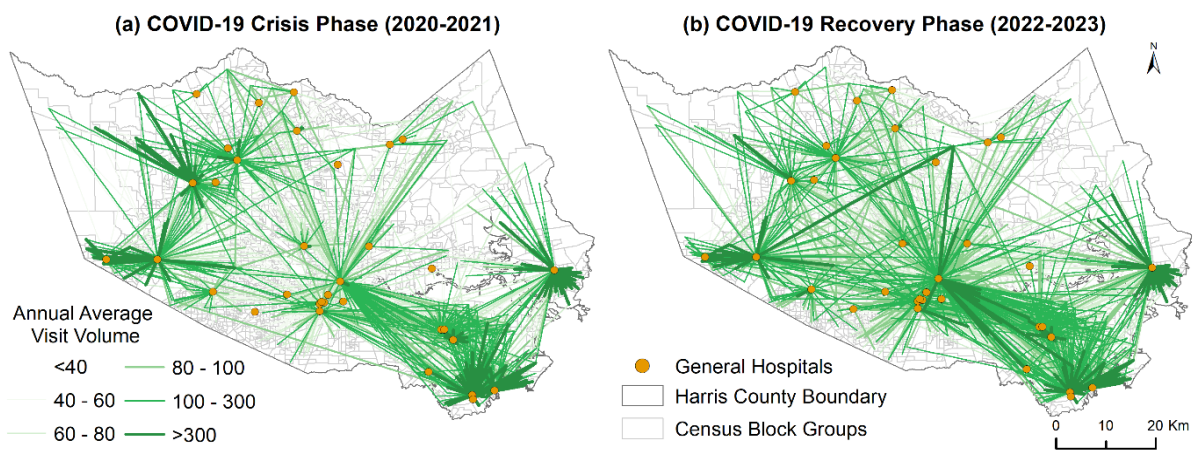

Figure 11 spatial patterns of annual average visits to general hospitals in Houston during the COVID-19 crisis phase (2020-2021) and COVID-19 recovery phase (2022-2023)

Third, this study has not yet applied the model to other cities for validation or comparative analysis. Although the findings offer valuable insights into the region under study, extending the analysis to diverse urban contexts could enhance the assessment of the model's generalizability. Comparative investigations across cities with differing demographic compositions, healthcare infrastructures, and geographic characteristics would provide a more comprehensive understanding of healthcare accessibility and further validate the robustness of the results.

Moreover, the current analysis focuses on overall spatial trends across the study area without explicitly examining intra-regional variations. However, healthcare accessibility and visitation behavior may differ substantially between urban cores and suburban or peripheral zones due to variations in population density, transportation connectivity, and facility distribution. Future studies could incorporate a spatial stratification analysis, for instance, comparing model performance and determinants between urban and suburban contexts, to evaluate the spatial heterogeneity and contextual consistency of the findings.

Finally, this study focused exclusively on General Acute Care Hospitals, which provide comparable inpatient and emergency services across the study area. While this design choice ensured analytical consistency, it also constrains the scope of generalization. Different hospital types, such as specialty care centers, rehabilitation facilities, and urgent care clinics, may attract distinct patient groups and exhibit different spatial interaction mechanisms (e.g., referral-based flows, chronic-care visits, or wait-time sensitivity). Future studies could extend this framework to multiple facility types or employ a hierarchical modeling approach to compare visitation dynamics across hospital categories.

## 6. Conclusion

Healthcare accessibility remains a critical issue, with significant disparities driven by SES, geographic distribution, and resource allocation. Existing research has often approached this problem in a fragmented manner, focusing on either hospital attributes or community demographics without accounting for their combined influence on visitation patterns. This study addresses this gap by integrating a comprehensive set of hospital and community-level factors, such as hospital capacities, quality ratings, and population SES, to analyze healthcare utilization patterns in Houston, Texas. Leveraging four years of granular human mobility data (2020–2023) from SafeGraph and social sensing data from Google Maps, this research employs five models—Naïve Regression, Gradient Boosting, MLPs, Deep Gravity, and HGNN—to predict healthcare visitation flows and uncover the complex interactions between distance, hospital attributes, and community demographics.

The findings provide critical insights into the factors shaping healthcare visitation spatial patterns. Hospital capacities, ICU occupancy rates, quality ratings, and popularity emerged as significant predictors, with their influence varying across travel distances. Convenience and stability are prioritized for shorter distances, while indicators of excellence, such as high ratings and resource availability, become more critical for longer distances. Population characteristics, including race, income, and education, further moderate these patterns, with higher visitation rates observed in areas with larger proportions of Hispanic, Black, under-18, and over-65 populations. White-majority areas show lower sensitivity to hospital ratings at shorter distances, while Asian populations and highly educated individuals prioritize quality, contrasting with less educated groups who emphasize geographic proximity. These findings offer valuable guidance for policymakers, highlighting the importance of improving hospital quality, expanding geographic accessibility, and tailoring interventions to meet the unique needs of underserved populations. By addressing these disparities, this research contributes to the development of more inclusive and sustainable healthcare systems and advances the understanding of complex healthcare accessibility dynamics.